%% file: main.tex
\documentclass[sigconf]{acmart}
\usepackage{algorithm}
\usepackage{algpseudocode}
\makeatletter
\def\BState{\State\hskip-\ALG@thistlm}
\makeatother
\usepackage{enumitem}
\usepackage{booktabs}
\usepackage{balance}

\copyrightyear{2019}
\acmYear{2019} 
\setcopyright{iw3c2w3}
\acmConference[WWW '19]{Proceedings of the 2019 World Wide Web Conference}{May 13--17, 2019}{San Francisco, CA, USA}
\acmBooktitle{Proceedings of the 2019 World Wide Web Conference (WWW '19), May 13--17, 2019, San Francisco, CA, USA}
\acmPrice{}
\acmDOI{10.1145/3308558.3313733}
\acmISBN{978-1-4503-6674-8/19/05}

\begin{document}

\title{CnGAN: Generative Adversarial Networks for Cross-network User Preference Generation for Non-overlapped Users}

\author{Dilruk Perera}
\affiliation{
	\institution{School of Computing}
	\institution{National University of Singapore}
}
\email{dilruk@comp.nus.edu.sg}

\author{Roger Zimmermann}
\affiliation{
	\institution{School of Computing}
	\institution{National University of Singapore}
}
\email{rogerz@comp.nus.edu.sg}

\begin{abstract} A major drawback of cross-network recommender solutions is that they can only be applied to users that are overlapped across networks. Thus, the non-overlapped users, which form the majority of users are ignored. As a solution, we propose CnGAN, a novel multi-task learning based, encoder-GAN-recommender architecture. The proposed model synthetically generates source network user preferences for non-overlapped users by learning the mapping from target to source network preference manifolds. The resultant user preferences are used in a Siamese network based neural recommender architecture. Furthermore, we propose a novel user-based pairwise loss function for recommendations using implicit interactions to better guide the generation process in the multi-task learning environment. We illustrate our solution by generating user preferences on the Twitter source network for recommendations on the YouTube target network. Extensive experiments show that the generated preferences can be used to improve recommendations for non-overlapped users. The resultant recommendations achieve superior performance compared to the state-of-the-art cross-network recommender solutions in terms of accuracy, novelty and diversity.
\end{abstract}



\maketitle

\input{samplebody-conf}

\bibliographystyle{ACM-Reference-Format}
\balance 
\bibliography{sample-bibliography}

\end{document}

%% file: samplebody-conf.tex
\section{Introduction}
Cross-network recommender systems utilize auxiliary information from multiple source networks to create comprehensive user profiles for recommendations on a target network. Therefore, unlike traditional recommender solutions which are limited to information within a single network, cross-network solutions are more robust against cold-start and data sparsity issues \cite{mehta2005ontologically}. For example, the additional information from Twitter and Flicker, increased recommender precision on Delicious by 10\% \cite{abel2011analyzing}. Similarly, the transfer of information from various source networks to target networks such as Google+ to YouTube \cite{deng2013personalized}, Twitter to YouTube \cite{roy2012socialtransfer, perera2017exploring} and Wikipedia to Twitter \cite{osborne2012bieber} have consistently improved recommender accuracy. Furthermore, cross-network solutions allow user preferences to be captured from diverse perspectives, which increases the overall recommender quality in terms of diversity and novelty \cite{perera2018lstm, perera2017exploring}. However, despite the growing success of cross-network recommender solutions, the majority of existing solutions can only be applied to users that exist in multiple networks (overlapped users). The remaining non-overlapped users, which form the majority are unable to enjoy the benefits of cross-network solutions.

Deep learning and generative modeling techniques have been successfully used in the recommender systems domain in the past few years. For example, restricted Boltzmann machines \cite{salakhutdinov2007restricted}, Autoencoders \cite{ouyang2014autoencoder}, Hidden Markov Models \cite{sahoo2012hidden} and Recurrent Neural Networks (RNN) \cite{hidasi2015session} are some of the models used for rating prediction tasks. Recent advancements in Generative Adversarial Networks (GANs) have also drawn recommender systems researchers to apply the new minimax game framework to information retrieval. For example, IRGAN \cite{wang2017irgan} and RecGAN \cite{bharadhwaj2018recgan} were designed to predict relevant documents to users. Existing solutions use GAN to generate simple rating values \cite{wang2017irgan,bharadhwaj2018recgan,yoo2017energy} or synthetic items (e.g., images \cite{kang2017visually}). In contrast, we use GAN to generate complex user data (i.e., user preferences on the source network).

We propose a novel GAN model (CnGAN) to generate cross-network user preferences for non-overlapped users. Unlike the standard GAN model, we view the learning process as a mapping from target to source network preference manifolds. The proposed model solves two main tasks. First, a generator task learns the corresponding mapping from target to source network user preferences and generates auxiliary preferences on the source network. Second, a recommender task uses the synthetically generated preferences to provide recommendations for users who only have interactions on the target network. Furthermore, we propose two novel loss functions to better guide the generator and recommender tasks. The proposed solution is used to provide recommendations for YouTube users by incorporating synthesized preferences on Twitter. However, the solution is generic and can be extended to incorporate multiple source networks.

In this paper, we first briefly provide preliminaries to the proposed model and present the proposed CnGAN solution detailing its two main components, the generator and recommender tasks. Then, we compare the proposed model against multiple baselines to demonstrate its effectiveness in terms of accuracy, novelty and diversity. We summarize our main contributions as follows:
\begin{itemize}[leftmargin=*]
\item {To the best of our knowledge, this is the first attempt to apply a GAN based model to generate missing source network preferences for non-overlapped users.}
\item {We propose CnGAN, a novel GAN based model which includes a novel content loss function and user-based pairwise loss function for the generator and recommender tasks.}
\item{We carry out extensive experiments to demonstrate the effectiveness of CnGAN to conduct recommendations for non-overlapped users and improve the overall quality of recommendations compared to state-of-the-art methods.}
\end{itemize}

\section{Model}
\label{mod}
\subsection{Model Preliminaries}
\label{modPre}
\subsubsection{Bayesian Personalized Ranking (BPR):}
\label{bayPerRan}
Learning user preferences from implicit feedback is challenging since implicit feedback does not indicate preference levels toward items, it only indicates the presence or absence of interactions with items. BPR \cite{rendle2009bpr} was proposed to rank items based on implicit feedback using a generic optimization criterion. BPR is trained using pairwise training instances $S = \{(u,i,j) |u \in U 	\land i \in I_{u+} 	\land j \in I \setminus I_{u+}\}$, where $I$ is the set of all items and $I_{u+}$ is the set of items user $u$ has interacted with. BPR uses a pairwise loss function to rank interacted items higher than non-interacted items, and maximizes their margin using a pairwise loss as follows:
\begin{equation}
\label{eq:one}
	L_{BPR}(S)= \sum_{(u,i,j) \in S} - \ln \sigma (\hat{r}_{uij}) + \lambda_\Theta {\| \Theta \|}^2
\end{equation}
\noindent where $\hat{r}_{uij} = \hat{r}_{ui} - \hat{r}_{uj}$, $\Theta$ are model parameters, $\sigma(x) = e^x/e^{x+1}$ is a sigmoid function and $\lambda_\Theta$ are regularization parameters. 


\subsubsection{Feature extraction on a continuous space:}
\label{extCon}
CnGAN is trained using target network interactions of each user at each time interval and their corresponding source network interactions. We used topic modeling to capture user interactions on a continuous topical space since CnGAN is a GAN based model and requires inputs in a continuous space. Let $u_o \in U = [Tn_u^t, Sn_u^t]$ denote an overlapped user at time $t$, where $Tn_u^t$ and $Sn_u^t$ are target and source network interactions spanned over $T = [1,\dots,t]$ time intervals. A non-overlapped user $u_{no} \in U = [Tn_u^t]$ is denoted only using target network interactions. We used YouTube as the target and Twitter as the source network to conduct video recommendations on YouTube. Therefore, $Tn_u^t$ is the set of interacted videos (i.e., liked or added to playlists) and $Sn_u^t$ is the set of tweets. We assumed that each interaction (video or tweet) is associated with multiple topics, and extracted the topics from the textual data associated with each interaction (video titles, descriptions and tweet contents). We used the Twitter-Latent Dirichlet Allocation (Twitter-LDA) \cite{zhao2011comparing} for topic modeling since it is most effective against short and noisy contents. Based on topic modeling, each user $u$ on the target network is represented as a collection of topical distributions $tn_u = \{\boldsymbol{tn_u^1}; \dots; \boldsymbol{tn_u^t}\} \in \mathbb{R}^{T \times K^t}$ over $T$ time intervals, where $K^t$ is the number of topics. Each vector $\boldsymbol{tn_u^t} = \{tn_u^{t,1}; \dots; tn_u^{t,K^t}\} \in \mathbb{R}^{K^t}$ is the topical distribution at time $t$, where $tn_u^{t,k} = \hat{tn}_u^{t,k}/ \sum_{c=1}^{K^t} \hat{tn}_u^{t,c}$ is the relative frequency of topic k, and $\hat{tn}_u^{t,k}$ is the absolute frequency of the corresponding topic. Similarly, on the source network, interactions are represented as $sn_u = \{\boldsymbol{sn_u^1}; \dots; \boldsymbol{sn_u^t}\} \in  \mathbb{R}^{T \times K^t}$. The resultant topical frequencies indicate user preference levels towards corresponding $K^t$ topics. Therefore, $u = \{tn_u, sn_u\} \in \mathbb{R}^{T \times 2K^t}$ represents user preferences over $T$ time intervals on a continuous topical space, and forms the input to the proposed model.

\subsection{Generator Task}
\label{cng}
The generator task learns the function that maps the target network preferences manifold to the source network preferences manifold using Encoders ($E$), Discriminator ($D$) and Generator ($G$).

\subsubsection{Encoders:}
\label{enc}
User preferences captured as topical distribution vectors $(\boldsymbol{tn_u^t}$ and $\boldsymbol{sn_u^t})$ become sparse when $\boldsymbol{K^t}$ is set to a sufficiently large value to capture finer level user preferences, and/or the number of interactions at a time interval is low. One of the most effective methods to train machine learning models with highly sparse data is to model interactions between input features \cite{blondel2016polynomial} (e.g., using neural networks \cite{he2017neural}). Therefore, we used two neural encoders $E_{tn}$ and $E_{sn}$ for target and source networks to transform the input topical distributions to dense latent encodings. The resultant encodings are represented as $E_{tn}(tn_u)$ $ \in \mathbb{R}^{T \times En}$ and $E_{sn}(sn_u)\in \mathbb{R}^{T \times En}$, where $En$ is the dimensionality of the latent space. These encodings form the input to the generator task.
\begin{figure}
	\centering
	\includegraphics[width=0.9\linewidth]{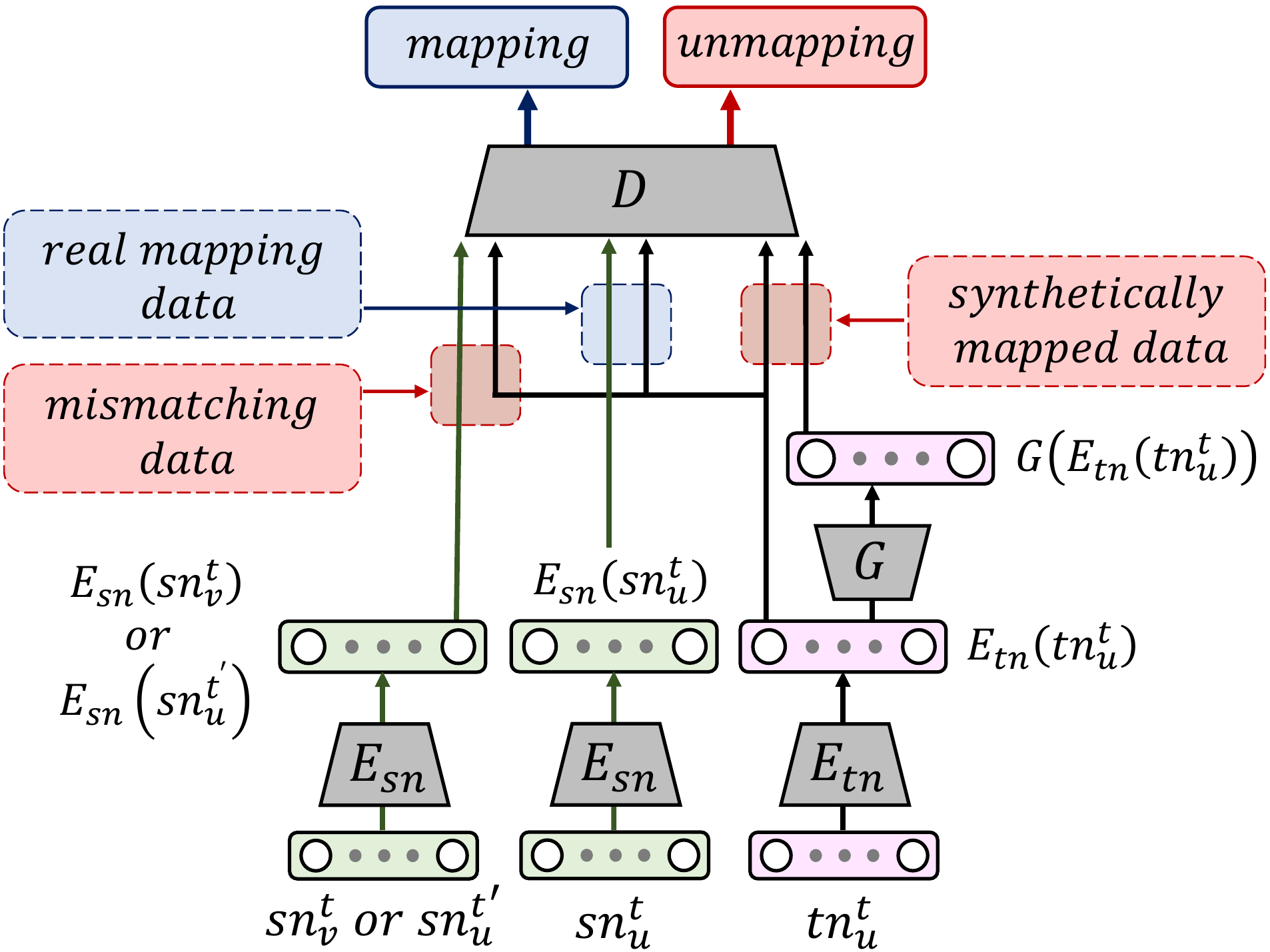}
	\caption{Adversarial learning process between $\boldsymbol{E}$, $\boldsymbol{D}$ and $\boldsymbol{G}$.}
	\label{fig:fig1}
\end{figure}

\subsubsection{Generator task formulation:} 
\label{genProbFor}
For a non-overlapped user $u$, let $E_{tn}(\boldsymbol{tn_u^t})$ denote the target network encoding at time interval $t$. The generator task aims to synthetically generate the mapping source network encoding that closely reflects the missing source network encoding $E_{sn}(\boldsymbol{sn_u^t})$. Hence, $G$ uses target encoding $E_{tn}(\boldsymbol{tn_u^t})$ and attempts to generate the mapping source encoding to fool $D$. Similarly, $D$ tries to differentiate between real source encoding $E_{sn}(\boldsymbol{sn_u^t})$ and generated encoding $G(E_{tn}(\boldsymbol{tn_u^t}))$. Note that, we refer to the actual and generated target and source network encodings of non-overlapped users $\big(E_{tn}(\boldsymbol{tn_u^t}), G(E_{tn}(\boldsymbol{tn_u^t}))\big)$ as \textit{synthetically mapped data} and the actual target and source network encodings of overlapped users $\big(E_{tn}(\boldsymbol{tn_u^t}), E_{sn}(\boldsymbol{sn_u^t})\big)$ as \textit{real mapping data}.

\subsubsection{Discriminator:} Analogous to $D$ in standard GANs, the real and synthetically mapped pairs could be used to learn $D$. However, $D$ may only learn to differentiate between actual and generated source network encodings in a given pair of inputs, without learning to check if the given pair is a mapping pair. Therefore, during training, we input mismatching source and target network encodings to ensure that an effective mapping is learned. We modified the loss function for $D$ to minimize output scores for mismatching pairs and maximize output scores for matching pairs. For a given target network encoding, we drew mismatching source encodings only from real data to avoid potential biases. Mismatching pairs were created by pairing real encodings from different users at the same time interval or by pairing encodings from the same user at different time intervals (see Figure \ref{fig:fig1}). Accordingly, the loss function for $D$ is formed as follows: 
\begin{equation}
\label{veed}
	\begin{array}{l}
		\max_{E_{tn},E_{sn},D} V(E_{tn},E_{sn},D)\\
        = \mathbb{E}_{\boldsymbol{{tn}_u^t},\boldsymbol{{sn}_u^t} \sim\ p_{data}(tn,sn)}L_{real}\Big(E_{tn}(\boldsymbol{tn_u^t}),E_{sn}(\boldsymbol{sn_u^t})\Big)\\
        + \mathbb{E}_{\boldsymbol{{tn}_u^t} \sim\ p_{data}(tn)}L_{fake}\Big(E_{tn}(\boldsymbol{tn_u^t}), G\Big(E_{tn}(\boldsymbol{tn_u^t})\Big)\Big)\\
        + \mathbb{E}_{\boldsymbol{{tn}_u^t},\boldsymbol{{\overline{sn}}_u^t} \sim\ \overline{p}_{data}(tn,\overline{sn})}L_{mismatch}\Big(E_{tn}(\boldsymbol{tn_u^t}),E_{sn}(\boldsymbol{\overline{sn}_u^t})\Big)       
        \end{array}
\end{equation}
\noindent where $p_{data}(tn,sn)$ and $\overline{p}_{data}(tn,\overline{sn})$ are matching and mismatching target and source network topical distributions, $p_{data}(tn)$ are the target network topical distributions and $G(x)$ is the generated matching source network encoding for the given target network encoding $x$. Furthermore, $L_{real}(x,y) = [\log(D(x,y))]$, $L_{fake}(x,y) = [1 - \log(D(x,y))]$ and $L_{mismatch}(x,y) = [1 - \log(D(x,y))]$, where $D(x,y)$ is the probability that $x$ and $y$ are matching target and source network encodings. Therefore, $D$ and $E$ work together to maximize the discriminator value function $V(E_{tn}, E_{sn}, D)$. The encoders ($E_{tn}$ and $E_{sn}$) assist the process by encoding the topical distributions that are optimized for learning $D$ (e.g., extract encodings that express the latent mapping between target and source networks).

\subsubsection{Generator:} Once $D$ and $E$ are trained, they adversarially guide $G$ by effectively identifying real and synthetically generated data. Specifically, $G$ takes in a target network encoding $E_{tn}(\boldsymbol{tn_u^t})$ and synthetically generates the mapping source network encoding that resemble $E_{sn}(\boldsymbol{sn_u^t})$ drawn from real mapping data. 

GAN models are typically used to generate real-like images from random noise vectors. For a given noise vector, different GANs can learn multiple successful mappings from the input space to the output space. In contrast, CnGAN aims to learn a specific mapping from target to source network encodings. Hence, for given pairs of real mapping data $(\boldsymbol{tn_u^t}, \boldsymbol{sn_u^t})$, $G$ minimizes an additional content loss between the generated $G(E_{tn}(\boldsymbol{tn_{u}^t}))$ and real $E_{sn}(\boldsymbol{sn_{u}^t})$ source encodings to provide additional guidance for $G$. Therefore, we formulated the loss function for $G$ as follows:
\begin{equation}
\label{vg}
	\begin{array}{l}
		\min_{G} V(G) = \mathbb{E}_{\boldsymbol{{tn}_u^t} \sim\ p_{data}(tn)}L_{fake}\Big(E_{tn}(\boldsymbol{tn_u^t}), G\Big(E_{tn}(\boldsymbol{tn_u^t})\Big) \Big)\\
        + \mathbb{E}_{\boldsymbol{{sn}_u^t}, \boldsymbol{{tn}_u^t} \sim\ p_{data}(tn,sn)}L_{content}\Big(E_{sn}(\boldsymbol{sn_u^t}), G\Big(E_{tn}(\boldsymbol{tn_u^t})\Big) \Big)       
        \end{array}
\end{equation}
\noindent where $L_{content}(x,y) = \|x,y \|_1$ is the content loss computed as the $\ell 1$-norm loss between real and generated encodings.

\subsection{Recommender Task}
\label{rec}
The recommender task uses the generated preferences to conduct recommendations for non-overlapped users.
\subsubsection{Recommender task formulation:} 
\label{recProbFor}
Let $E_{tn}(\boldsymbol{tn_{u}^t}), G(E_{tn}(\boldsymbol{tn_{u}^t}))$ denote a non-overlapped user $u$ at time $t$ based on synthetically mapped data, which reflects current preferences. To capture previous preferences, we used a previous interaction vector $ R_{u}^{t-}\in \mathbb{R}^M$, where $M$ is the number of items. Each element $r_{{u},i}^{t-} \in \boldsymbol{R_{u}^{t-}}$ is set to 1 if the user had an interaction with item $i$ before time $t$, and 0 if otherwise. Thus, we formulated the recommender task as a time aware Top-N ranking task where, current and previous user preferences are used to predict a set of $N$ items that the user is most likely interact with, in the next time interval $t+1$.

\subsubsection{Recommender loss:} 
\label{recLos}
Typically, BPR optimization is performed using Stochastic Gradient Descent (SGD) where, for each training instance $(u,i,j)$, parameter ($\Theta$) updates are performed as follows:
\begin{equation}
\label{eq:two}
\begin{gathered}
	\Theta \leftarrow  \Theta - \alpha \frac{\partial}{\partial \Theta} L_{BPR}(S) \\
	\Theta \leftarrow \Theta + \alpha \bigg(\frac{e^{-\hat{r}_{uij}}}{1+e^{-\hat{r}_{uij}}} \cdot \frac{\partial}{\partial \Theta} \hat{r}_{uij} + \lambda_\Theta \Theta \bigg)
\end{gathered}
\end{equation}
where $\alpha$ is the learning rate, $\hat{r}_{uij} = \hat{r}_{ui} - \hat{r}_{uj}$ is the pairwise loss for a given $(u, i, j)$ triplet and $\hat{r}_{ui}$ is the predicted rating for user $u$ and item $i$. Since BPR is a generic optimization criterion, $\hat{r}_{ui}$ can be obtained using any standard collaborative filtering approach (e.g., MF). Considering the rating prediction function in MF ($\hat{r}_{ui} = \sum_{f=1}^K \boldsymbol{w_{uf}} \cdot \boldsymbol{h_{if}}$, where $\boldsymbol{w_{uf}}\in \mathbb{R}^K$ and $\boldsymbol{h_{if}}\in \mathbb{R}^K$ are latent representations of user $u$ and item $i$), $\hat{r}_{uij}$ for MF based BPR optimization can be defined as $\hat{r}_{uij} = \sum_{f=1}^K w_{uf} \cdot (h_{if} - h_{jf})$.

For each $(u, i, j)$ instance, three updates are performed when $\Theta = w_{uf}$, $\Theta = h_{if}$ and $\Theta = h_{jf}$. For each update on the latent user representation ($w_{uf}$), two updates are performed on the latent item representations $(h_{if}$ and $h_{jf})$. Hence, the training process is biased towards item representation learning, and we term the standard BPR optimization function as an \textit{item-based} pairwise loss function. Since our goal is to learn effective user representations to conduct quality recommendations, we introduce a \textit{user-based} pairwise loss function ($R$ loss), which is biased towards user representation learning. 

We used a fixed time-length sliding window approach in the training process where, at each time interval, the model is trained using the interactions within the time interval (see Section \ref{sub:model_offline_train}). Thus, we denote the training instances for the new user-based BPR at each time interval $t$ as $S^t = \{(u,v,i)|u \in U_{i+}^t \land v \in U \setminus U_{i+}^t \land i \in I \}$ where $U_{i+}^t$ is the set of users who have interactions with item $i$, at time $t$. Compared to $(u,i,j)$ in the standard item-based BPR, user-based BPR contains $(u,v,i)$ where, for item $i$ at time $t$, user $u$ has an interaction and user $v$ does not. Thus, intuitively, the predictor should assign a higher score for $(u,i)$ compared to $(v,i)$. Accordingly, we defined the user-based loss function to be minimized as follows: 
\begin{equation}
	L_{UBPR}(S^t)= \sum_{(u,v,i) \in S^t} - \ln \sigma (\hat{r}_{uvi}^t) + \lambda_\Theta {\| \Theta \|}^2
\end{equation}
where $\hat{r}_{uvi}^t = \hat{r}_{ui}^t - \hat{r}_{vi}^t = \sum_{f=1}^{K} (w_{uf}^t - w_{vf}^t) h_{if}$. During optimization, $\Theta$ updates are performed as follows: \begin{equation}
	\Theta \leftarrow \Theta + \alpha \bigg(\frac{e^{-\hat{r}_{uvj}^t}}{1+e^{-\hat{r}_{uvj}^t}} \cdot \frac{\partial}{\partial \Theta} \hat{r}_{uvj}^t + \lambda_\Theta \Theta \bigg) \\
\end{equation}
For different $\Theta$ values, the following partial derivations are obtained:
\begin{equation}
    \frac{\partial}{\partial \Theta} \hat{r}_{uvj}^t = 
    \begin{cases}
    h_{if},  \text{if } \Theta = w_{uf}^t  \\
    -h_{if}, \text{if } \space \Theta = w_{vf}^t \\
    (w_{uf}^t - w_{vf}^t), \text{if } \space \Theta = h_{if} \\
    0, \text{otherwise}
\end{cases}
\end{equation}

\noindent Since the user-based BPR performs two user representation updates for each training instance, it is able to efficiently guide $G$ to learn user representations during training.

\subsubsection{Recommender architecture:}
\label{recMod}
\begin{figure}
	\centering
	\includegraphics[width=\linewidth]{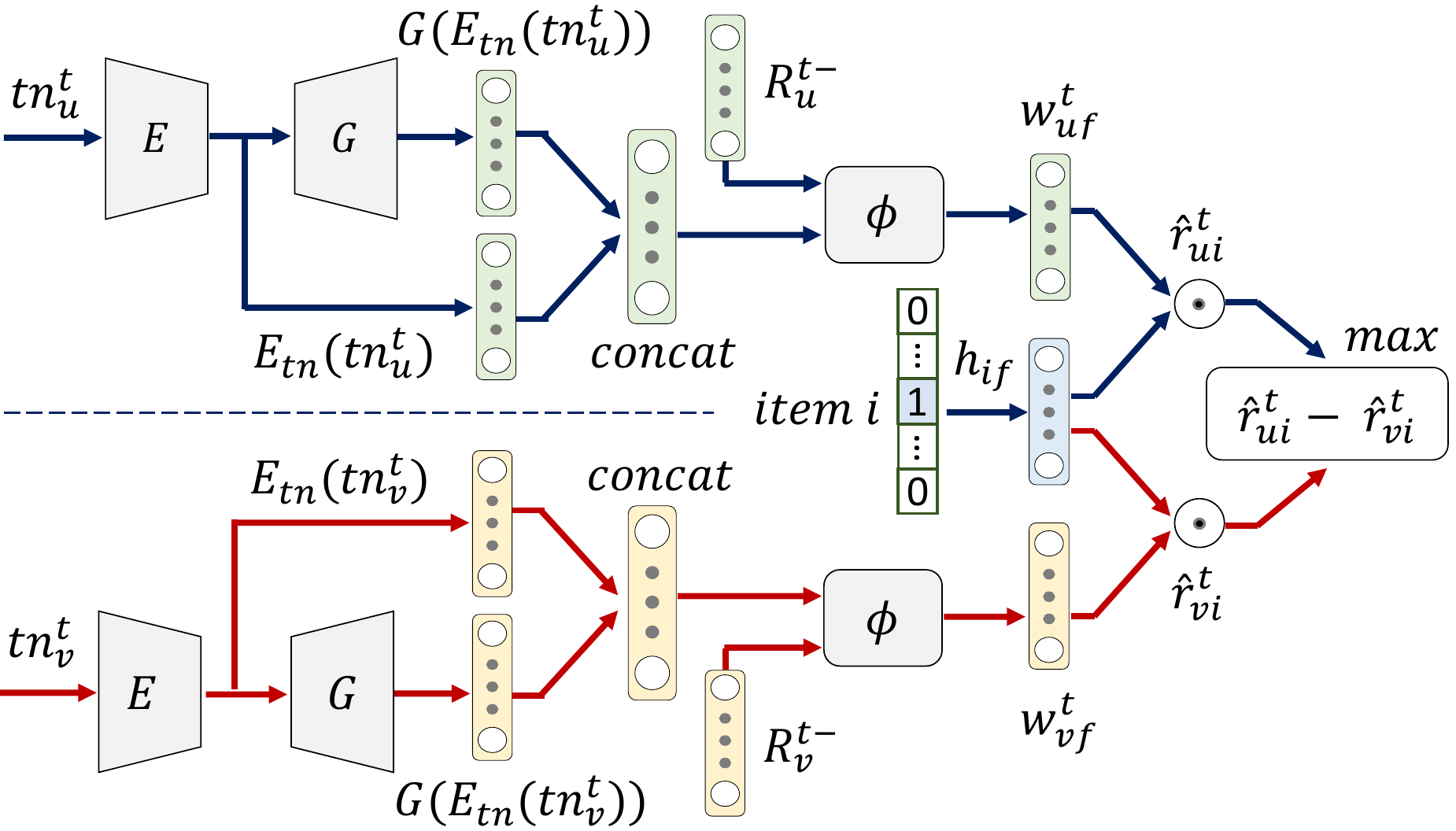}
	\caption{Recommender task architecture as a Siamese network.}
	\label{fig:fig2}
\end{figure}

We used a Siamese network for the recommender architecture since it naturally supports pairwise learning (see Figure \ref{fig:fig2}). Given a non-overlapped user $u$ with current preferences $E_{tn}(\boldsymbol{tn_u^t}), G(E_{tn}(\boldsymbol{tn_u^t}))$ and previous preferences $\boldsymbol{R_u^{t-}}$ at time $t$, we learned a transfer function $\Phi$, which maps the user to the latent user space $w_{u,f}^t$ for recommendations as follows:
\begin{equation}
\label{wuft}
	w_{u,f}^t = \Phi (E_{tn}(\boldsymbol{tn_u^t}), G(E_{tn}(\boldsymbol{tn_i^t})), \boldsymbol{R_u^{t-}})
\end{equation}

The transfer function $\Phi$ is learned using a neural network since neural networks are more expressive and effective than simple linear models. Accordingly, for a non-overlapped user $u$ at time $t$, the predicted rating $\hat{r}_{ui}$ for any given item $i$ is obtained using the inner-product between the latent user and item representations as follows:
\begin{equation}
\label{eq:noov_rating_pred}
	\hat{r}_{ui}^t = \sum_{f=1}^K \Phi (E_{tn}(tn_u^t), G(E_{tn}(tn_u^t)), \boldsymbol{R_u^{t-}}) \cdot \boldsymbol{h_{if}}	
\end{equation} 

\subsection{Multi-Task Learning (MTL)}
\label{multalea}
Although the generator and recommender tasks are separate, they depend on each other to achieve a higher overall performance. The generator task increases the accuracy of the recommender by synthesizing source network encodings that effectively represent user preferences. The recommender task guides the generator to efficiently learn the target to source mapping by reducing the search space. Therefore, we trained both interrelated tasks in a MTL environment to benefit from the training signals of each other. Hence, we formulated the multi-task training objectives as follows:

\begin{equation}
\min_{G} \max_{E_{tn}, E_{sn}, D} V(E_{tn}, E_{sn}, G, D) \\
\end{equation}
\begin{equation}
\label{lubpr}
\min_{w_{uf}^t,w_{vf}^t,h_{if}}L_{UBPR}(S^t)
\end{equation}
where $V(E_{tn},E_{sn},G,D)$ is the value function for $E$, $G$ and $D$, $L_{UBPR}(S^t)$ is the $R$ loss function, and $w_{uf}^t, w_{vf}^t$ and $h_{if}$ are model parameters. The $R$ loss is back-propagated all the way to $G$ since $w_{uf}^t$ and $w_{vf}^t$ are compositions of the functions $G$ and $\Phi$ (see Equation \ref{wuft}) and the generator and recommender tasks are trained as an end-to-end process. Hence, equation \ref{lubpr} can be restated as follows:
\begin{equation}
\min_{\Phi,G,h_{if}}L_{UBPR}(S^t)
\end{equation}

\section{Experiments}
\subsection{Dataset}
Due to the lack of publicly available timestamped cross-network datasets, we extracted overlapped users on YouTube (target) and Twitter (source) networks from two publicly available datasets \cite{yan2014mining,lim2015mytweet}. We scraped timestamps of interactions and associated textual contents (video titles, descriptions and tweet contents) over a 2-year period from 1\textsuperscript{st} March 2015 to 29\textsuperscript{th} February 2017. In line with common practices, we filtered out users with less than 10 interactions on both networks for effective evaluations. The final dataset contained 2372 users and 12,782 YouTube videos.

\subsection{Experimental Setup}
The recommender systems literature often uses training and testing datasets with temporal overlaps \cite{campos2012performance}, which provides undue advantages as it does not reflect a realistic recommender environment. To avoid such biases, at each time interval, the proposed model is trained using only the interactions during the current and previous time intervals. We randomly selected 50\% of users as non-overlapped users and removed their source network interactions. The model was first trained offline using overlapped users (see Section \ref{sub:model_offline_train}). Then, testing and online training were conducted using both overlapped and non-overlapped users (see Section \ref{sub:model_test}).

\subsubsection{Offline Training:}
\label{sub:model_offline_train}
We used the sliding window approach and data from the first 16 months ($2T/3$) of overlapped users to learn the mapping from target to source network preferences for the generator task, the neural transfer function $(\Phi)$, and item latent representations $(h_{if})$ for the recommender task. At each training epoch, the generator task is learned first. Hence, $E$, $D$ and $G$ are trained using real mapping data and the trained $G$ is used to generate synthetically mapped data. The generated preferences are used as inputs to the recommender task to predict interactions at the next time interval. The ground truth interactions at the same time interval are used to propagate the recommender error from the recommender to the generator task and learn parameters of both processes.

\subsubsection{Testing and Online Training:} 
\label{sub:model_test}
We used data from the last 8 months ($2T/3$ onward) to test and retrain the model online in a simulated real-world environment. At each testing time interval $t$, first, mapping source network preferences are generated for non-overlapped users based on their target network preferences. Second, the synthetically mapped data for non-overlapped users are used as inputs to conduct Top-N recommendations for $t+1$ (see Equation \ref{eq:noov_rating_pred}). The entire model is then retrained online before moving to the next time interval, as follows:  First, the recommender parameters $(\Phi$ and $\boldsymbol{h_{if}^t})$ are updated based on the recommendations for both overlapped and non-overlapped users. Second, the components of the generator task $(E$, $D$ and $G)$ are updated based on the real mapping data from overlapped users. Finally, to guide $G$ from the training signals of the recommender, the model synthesizes mapping data for overlapped users, which are used as inputs for recommendations. The error is back-propagated from the recommender task to $G$, and consequently, the parameters $G$, $\Phi$ and $\boldsymbol{h_{if}^t}$ are updated. This process is repeated at subsequent time intervals as the model continues the online retraining process. In line with common practices in sliding window approaches, the model is retrained multiple times before moving to the next time interval.

\subsection{Evaluation}
We formulated video recommendation as a Top-N recommender task and predicted a ranked set of $N$ videos that the user is most likely to interact with at each time interval. To evaluate model accuracy, we calculated the Hit Ratio (HR) and Normalized Discounted Cumulative Gain (NDCG)  \cite{he2015trirank}. Both metrics were calculated for each participating user at each time interval and the results were averaged across all users and testing time intervals.

\subsubsection{Baselines}
We evaluated CnGAN against single network, cross-network, linear factorization and GAN based baselines.
\begin{itemize}[leftmargin=*]
	\setlength\itemsep{0em}
	\item \textbf{TimePop}: Recommends the most popular $N$ items in the previous time interval to all users in the current time interval. 
    \item \textbf{TBKNN} \cite{campos2010simple}: Time-Biased KNN computes a set of $K$ neighbors for each user at each time interval based on complete user interaction histories, and recommends their latest interactions to the target user at the next time interval. Similar to the original work, we used several $K$ values from 4 to 50, and the results were averaged.
	\item \textbf{TDCN} \cite{perera2017exploring}: Time Dependent Cross-Network is an offline, MF based cross-network recommender solution, which provides recommendations only for overlapped users. TDCN learns network level transfer matrices to transfer and integrate user preferences across networks and conduct MF based recommendations.
	\item \textbf{CRGAN}: Due to the absence of GAN based cross-network user preference synthesizing solutions, we created Cross-network Recommender GAN, as a variation of our solution. Essentially, CRGAN uses the standard $D$ and $G$ loss functions and does not contain the $L_{mismatch}$ and $L_{content}$ loss components.
	\item \textbf{NUBPR-O} and \textbf{NUBPR-NO}: Due to the absence of neural network based cross-network recommender solutions, we created two Neural User-based BPR solutions, as variations of our solution. Essentially, both models do not contain the generator task. NUBPR-O considers both source and target network interactions and NUBPR-NO only considers target network interactions to conduct recommendations.
\end{itemize}

\subsection{Model Parameters}
\label{modPar}
We used Adaptive Moment Estimation (Adam) \cite{kingma2014adam} for optimizations since Adam adaptively updates the learning rate during training. We set the initial learning rate to 0.1, a fairly large value for faster initial learning before the rate is updated by Adam. We used only one hidden layer for all neural architectures, and given the size of the output layer $H_L$, the size of the hidden layer was set to $H_L\times 2$ to reduce the number of hyper-parameters. We used the dropout regularization technique and the dropout was set to 0.4 during training to prevent neural networks from overfitting.

\section{Discussion}  

\subsubsection{Offline training loss:}
\begin{figure}[!h]
\begin{minipage}[t]{0.45\linewidth}
    \centering
    \includegraphics[width=0.8\textwidth]{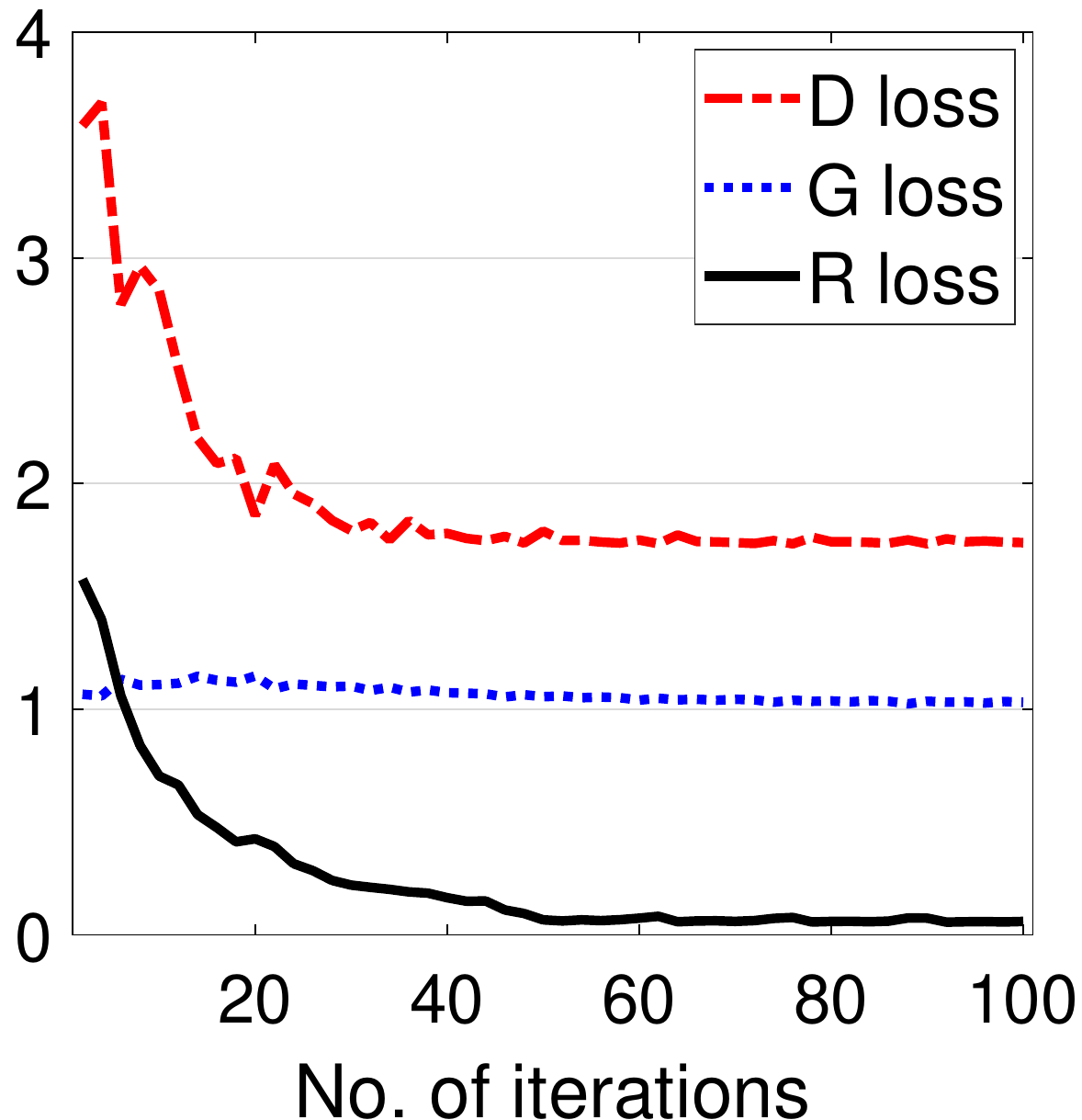}
    \caption{Offline training loss of $\boldsymbol{D}$, $\boldsymbol{G}$ and $\boldsymbol{R}$.}
    \label{fig:fig3}
\end{minipage}
\hspace{0.1cm}
\begin{minipage}[t]{0.48\linewidth} 
    \centering
    \includegraphics[width=0.8\textwidth]{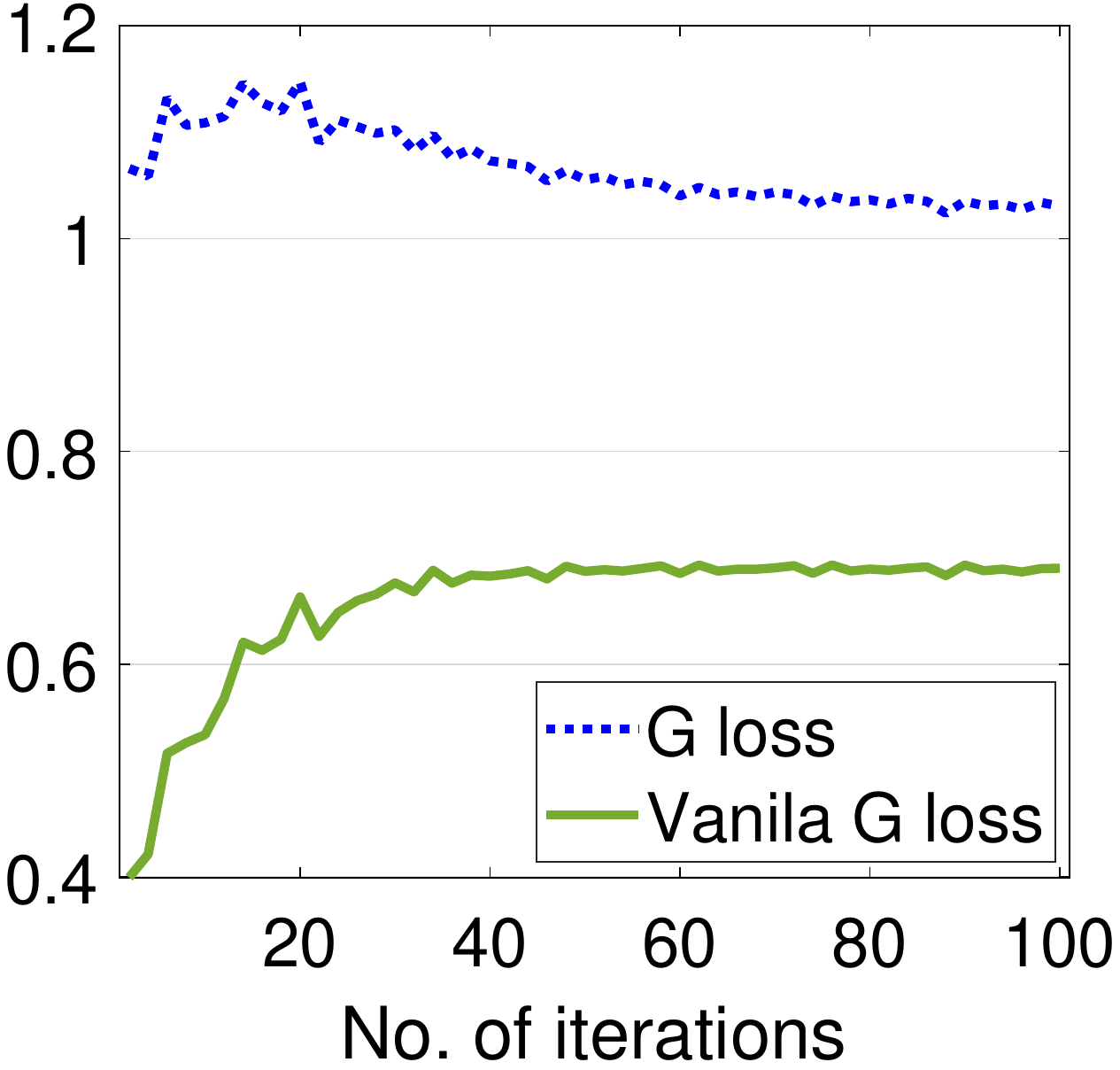}
    \caption{Offline training loss of $\boldsymbol{G}$ and vanilla $\boldsymbol{G}$.}
    \label{fig:fig4}
\end{minipage}        
\end{figure}

We compared the offline training loss of $D$, $G$ and $R$ (see Figure \ref{fig:fig3}). All three processes converge around 50 epochs, and compared to $D$ and $R$, the drop in $G$ is low. Therefore, we further examined the proposed $G$ loss against the vanilla $G$ loss for the same training process (see Figure \ref{fig:fig4}). Inline with standard GANs, the vanilla $G$ loss component in the proposed $G$ loss increases while the proposed $G$ loss slowly decreases. Hence, despite the low decrease in the overall $G$ loss, $L_{content}$ loss has notably decreased.

\subsubsection{Online training loss:}
\label{onTrainLoss}
\begin{figure}
	\centering
	\includegraphics[width=0.4\linewidth]{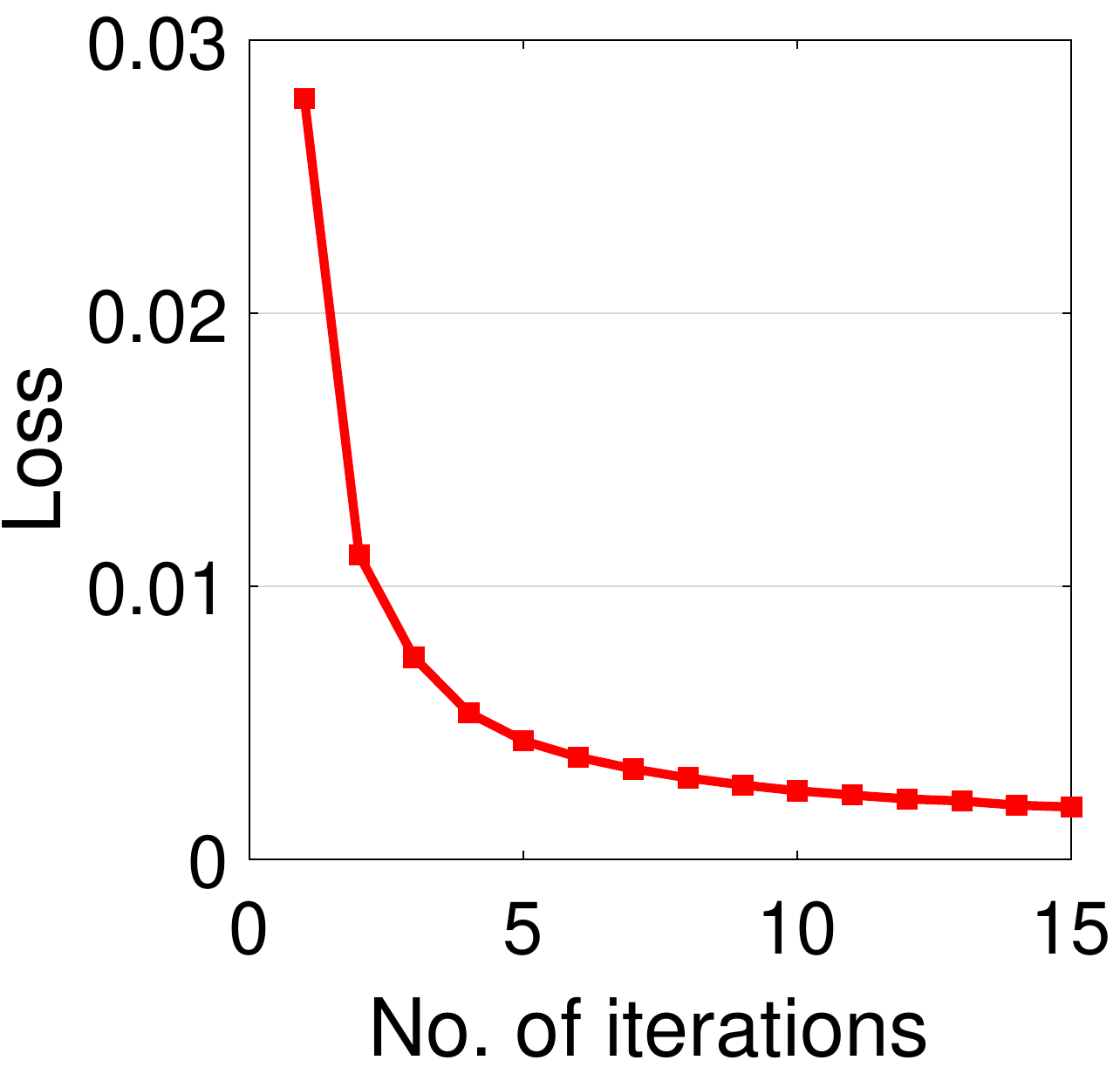}
	\caption{Online training loss of the recommender task.}
	\label{fig:fig5}
\end{figure}
We observed $R$ loss during the online training process (see Figure \ref{fig:fig5}). Updates are most effective within the first 10-15 iterations and after around 20 iterations, the recommender accuracy reaches its peak. Additional iterations tend to overfit the model and degrade performance. Therefore, the proposed solution is feasible in real-world applications since online training requires only a few iterations and retraining is costly.
\subsubsection{Prediction accuracy:}
\begin{figure}[!h]
\begin{minipage}[t]{0.48\linewidth}
    \centering
    \includegraphics[width=1\textwidth]{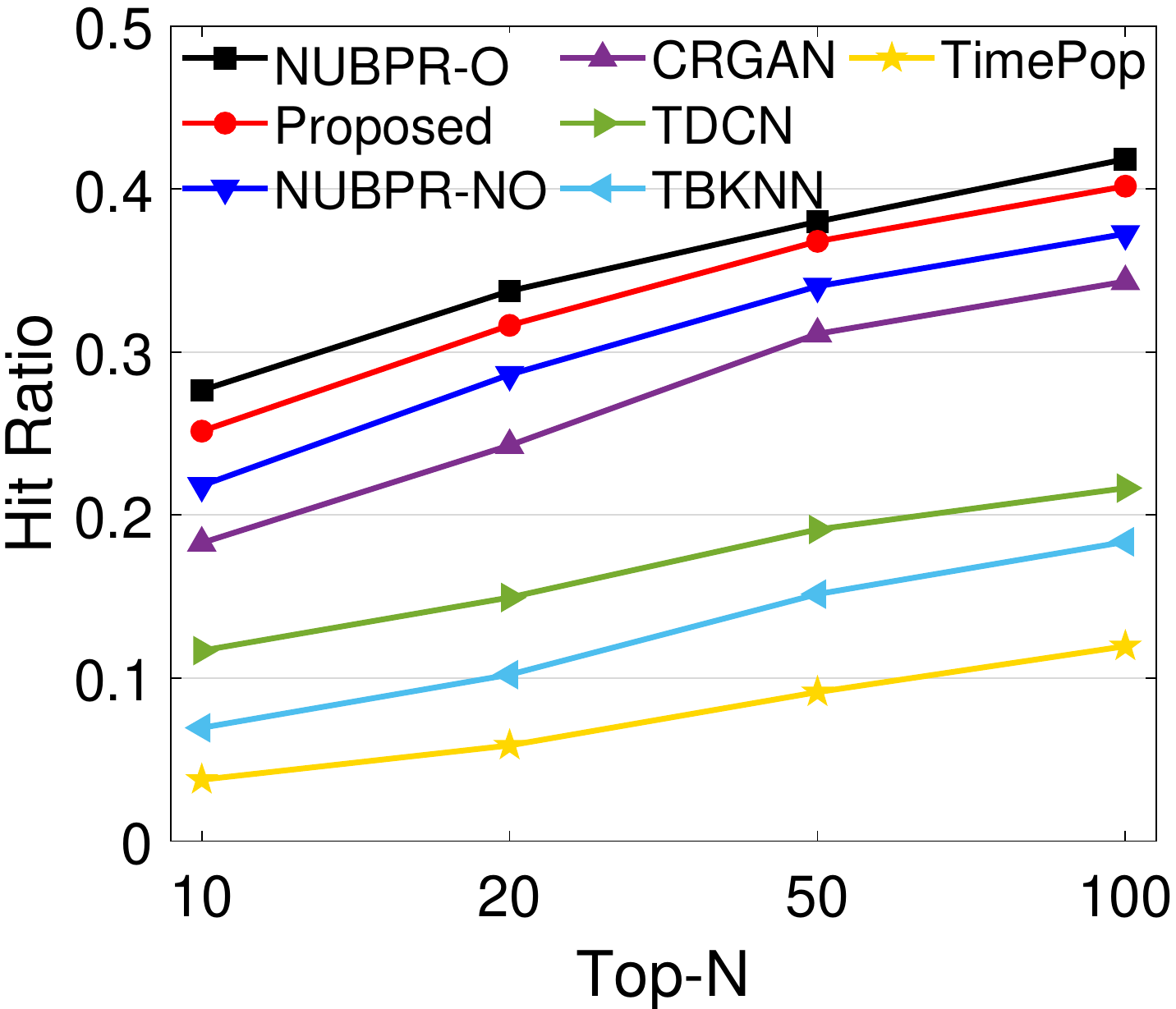}
\end{minipage}
\hspace{0.1cm}
\begin{minipage}[t]{0.48\linewidth} 
    \centering
    \includegraphics[width=1\textwidth]{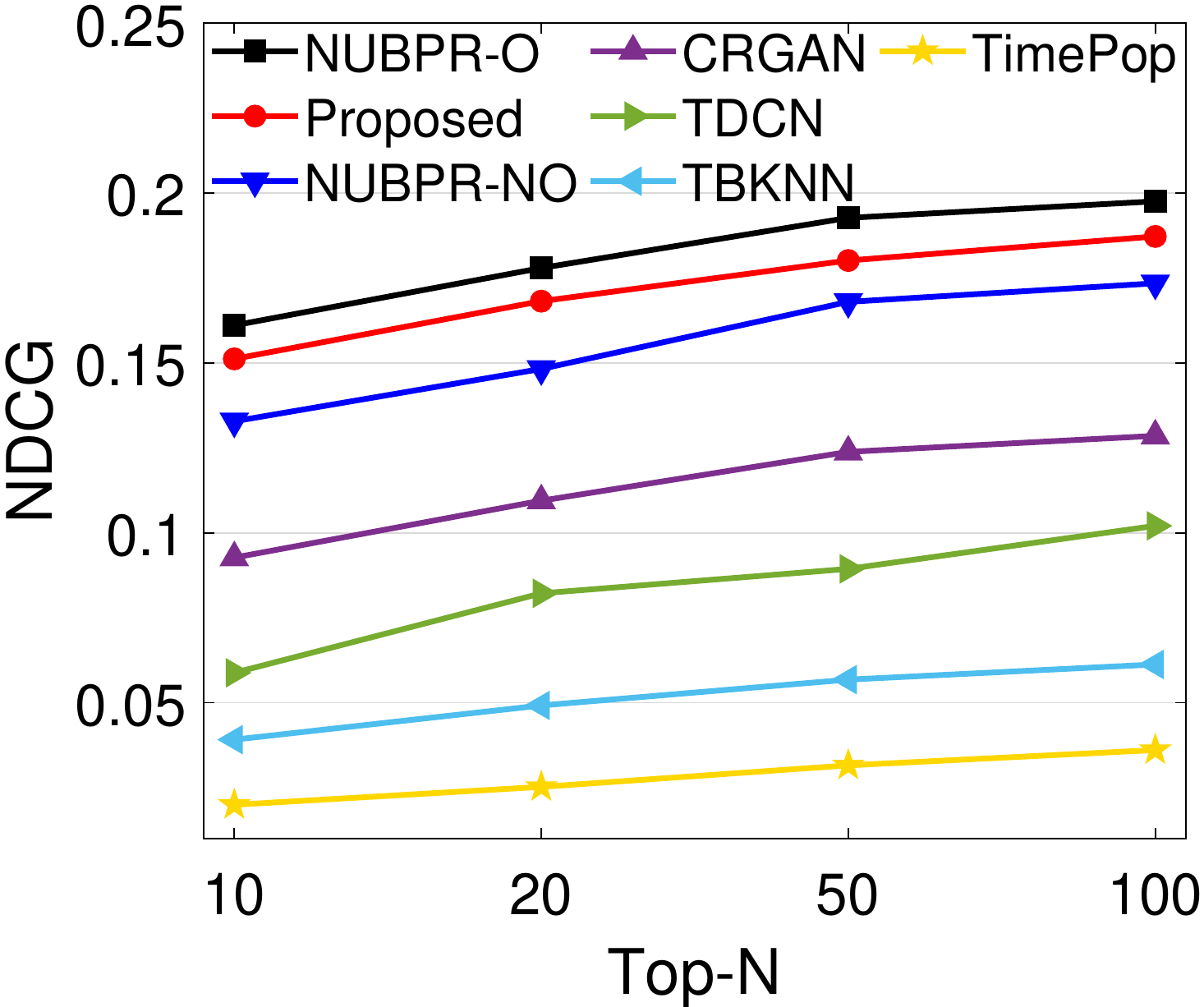}
\end{minipage}
\caption{Hit Ratio and NDCG for different Top-N values.}
\label{fig:fig6}
\end{figure}
We compared recommender accuracy (HR and NDCG) against multiple Top-N values (see Figure \ref{fig:fig6}). The neural network based solutions (Proposed, NUBPR-O, NUBPR-NO and CRGAN) show a higher accuracy since they capture complex relationships in user preferences. Among the non-neural network based solutions, TimePop has the lowest accuracy since it is based on a simple statistic - the popularity of videos. TDCN outperforms TBKNN and TimePop since it effectively utilizes both source and target network interactions to model user preferences.

As expected, NUBPR-O which does not have a data generation process and is only trained and tested with real mapping data from overlapped users has the best accuracy. We used NUBPR-O as the benchmark, since the goal is to synthesize data similar to the data used to train NUBPR-O. The proposed model shows the closest accuracy to NUBPR-O and consistently outperforms NUBPR-NO, which only used target network interactions. Therefore, the proposed model is able to generate user preferences that increases the recommender accuracy, even when the user overlaps are unknown. Furthermore, the improvements gained over CRGAN, which uses vanilla GAN loss functions, show the effectiveness of the proposed $D$ and $G$ loss functions.

\subsubsection{Prediction accuracy for different number of topics ($K^t$):}
\begin{figure}[!h]
\begin{minipage}[t]{0.48\linewidth}
    \centering
    \includegraphics[width=1\textwidth]{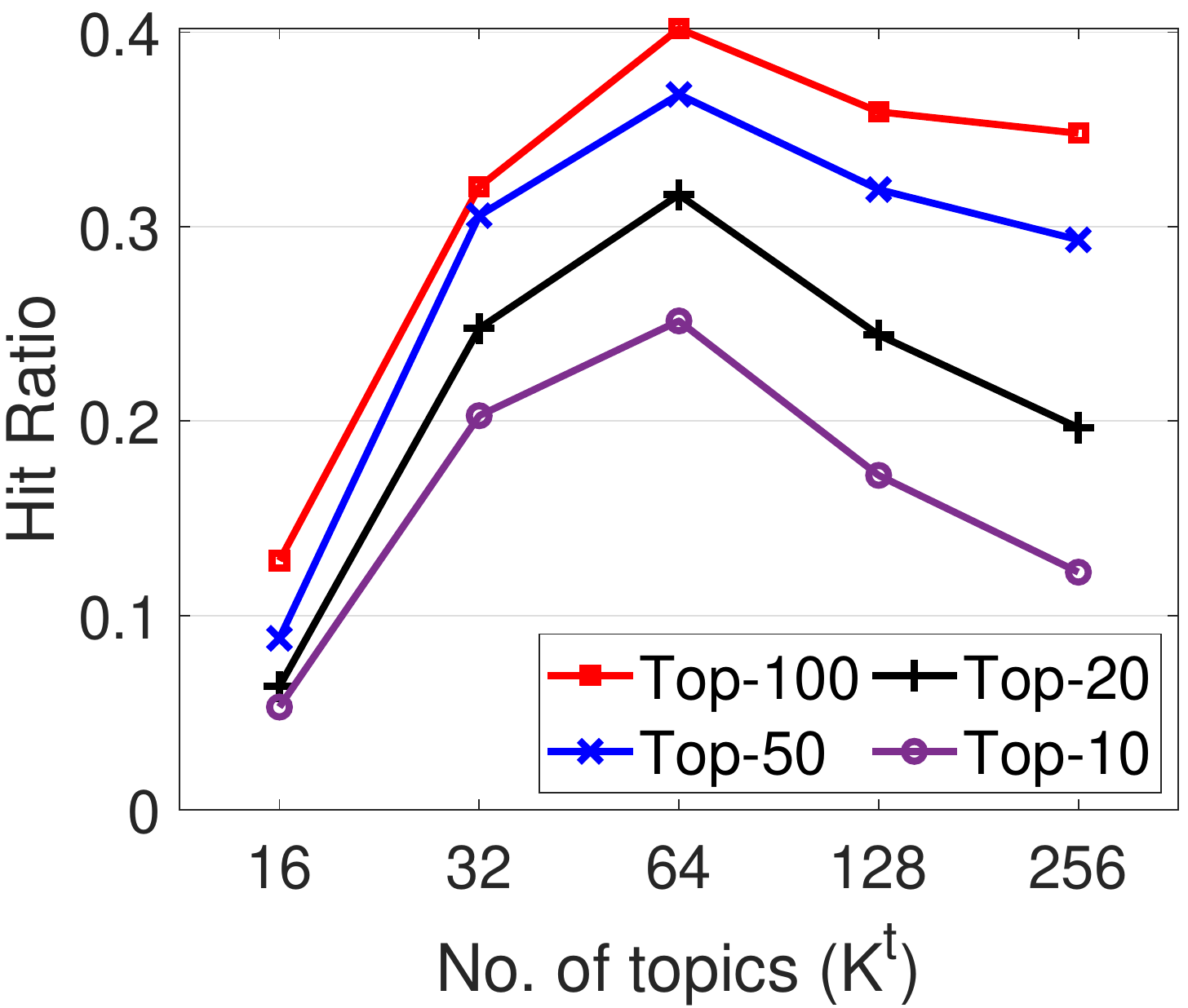}
\end{minipage}
\hspace{0.1cm}
\begin{minipage}[t]{0.48\linewidth} 
    \centering
    \includegraphics[width=1\textwidth]{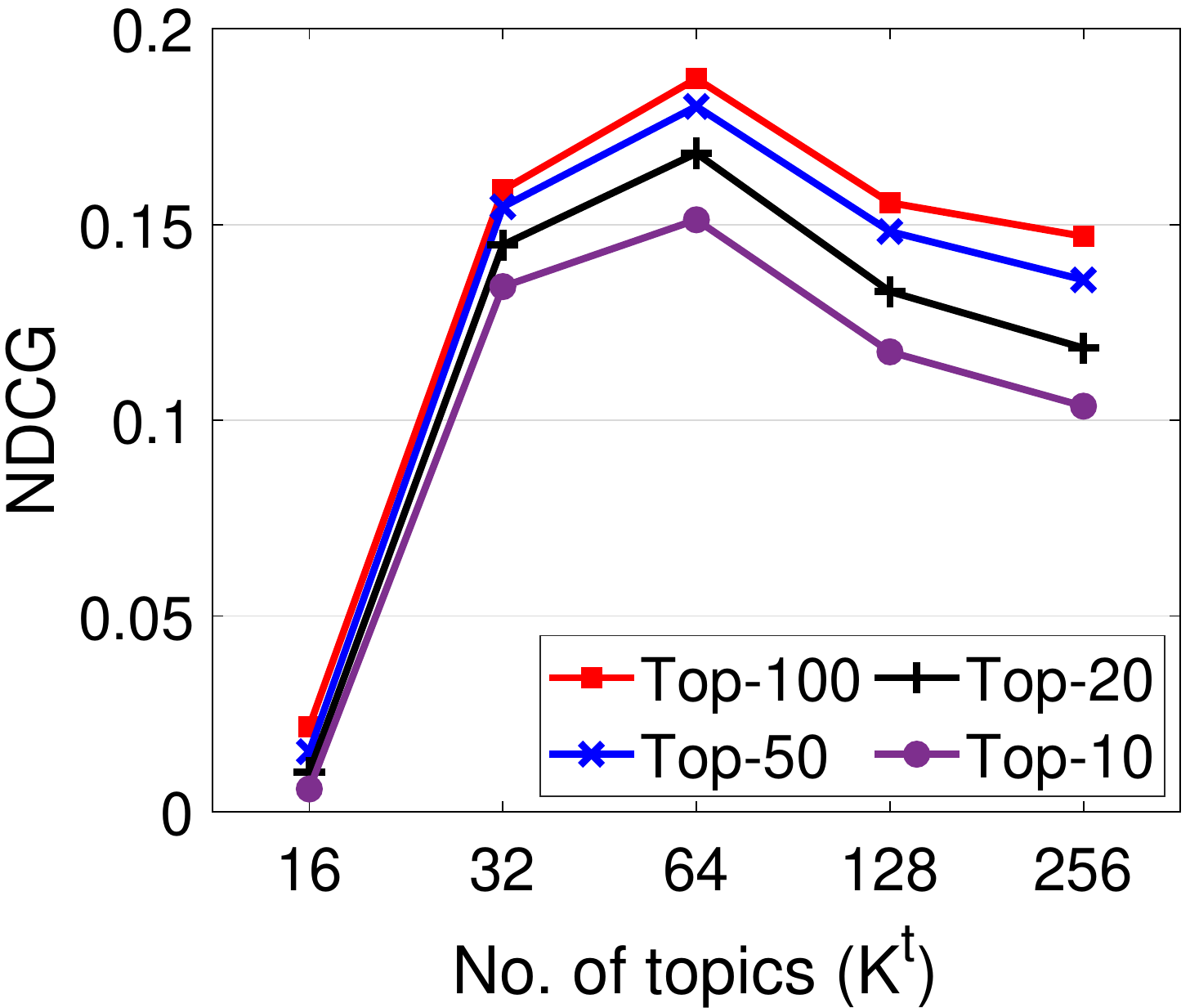}
\end{minipage}
\caption{Hit Ratio and NDCG for different number of topics.}
\label{fig:fig7}
\end{figure}

We compared recomender accuracy against the dimensionality  of the encoded topical space ($K^t$) (see Section \ref{extCon} and Figure \ref{fig:fig7}). A grid search algorithm was used to select an effective $K^t$ value, and the highest accuracy is achieved when $K^t$ is around 64 topics for all Top-N values. Smaller number of topics fail to effectively encode diverse user preferences, while a higher number of topics create highly sparse inputs and reduce recommender performance. Further experiments using the standard perplexity measure \cite{blei2003latent} showed that 64 topics lead to a smaller perplexity with faster convergence.

\subsubsection{Diversity and novelty:} A higher recommendation accuracy alone is insufficient to capture overall user satisfaction. For example, continuously recommending similar items (in terms of topics, genre, etc.) could lead to a decline in  recommendation accuracy as users lose interest over time. Therefore, we compared the diversity \cite{avazpour2014dimensions} and novelty \cite{zhang2013definition} of recommended items. The proposed model was able to recommend videos with better novelty and diversity, and was the closest to NUBPR-O model (only a 2.4\% and 3.8\% novelty and diversity drop). Compared to the single-network based solutions, cross-network solutions showed better results as they utilize rich user profiles with diverse user preferences. Against the closest CRGAN approach, the proposed model showed considerable improvements in novelty (by 8.9\%) and diversity (by 12.3\%).

\subsubsection{Alternative GAN architectures:} Designing and training GAN models can be challenging due to the unbalance between $D$ and $G$, diminishing gradients and non-convergence \cite{arjovsky2017towards, salimans2016improved}. Various GAN architectures and training techniques were recently introduced to handle such issues. Therefore, despite the effectiveness of CnGAN, we designed two alternative solutions based on two widely popular GAN architectures, Deep convolutional GAN (DCGAN) \cite{radford2015unsupervised} and Wasserstein GAN (WGAN) \cite{arjovsky2017wasserstein} to replace the $G$ and $D$ learning processes. We found that DCGAN becomes highly unstable in this environment, where D error is constantly increased. This can be because, DCGAN is based on Convolution Neural Networks, known to capture the local features within the input data. However, unlike typical image data, user preferences in the vectors may not provide any interesting local features. Hence, DCGAN was less effective. Further experiments using WGAN also did not improve the performances.

\section{Conclusion and Further Work}
\label{conFurWor}
Typical cross-network recommender solutions are applied to users that are fully overlapped across multiple networks. Thus to the best of our knowledge, we propose the first Cross-network Generative Adversarial Network based model (CnGAN), which generates user preferences for non-overlapped users. The proposed model first uses a generator task to learn the mapping from target to source network user preferences and synthesize source network preferences for non-overlapped users. Second, the model uses a recommender task based on a Siamese network to incorporate synthesized source network preferences and conduct recommendations. The proposed model consistently outperformed multiple baselines in terms of accuracy, diversity and novelty of recommendations. As future work, we plan to investigate the recommender quality improvements when using both generated and real data for overlapped users. Furthermore, the model can be extended to use social information of the users. Overall, CnGAN alleviates a significant limitation in cross-network recommender solutions, and provides a foundation to make quality cross-network recommendations for all users. 

\subsection*{Acknowledgments}
This research has been supported by Singapore Ministry of Education Academic Research Fund Tier 2 under MOE's official grant number MOE2018-T2-1-103. We gratefully acknowledge the support of NVIDIA Corporation with the donation of a Titan Xp GPU used for this research.
\newpage

%% file: main.bbl

\begin{thebibliography}{32}


\ifx \showCODEN    \undefined \def \showCODEN     #1{\unskip}     \fi
\ifx \showDOI      \undefined \def \showDOI       #1{#1}\fi
\ifx \showISBNx    \undefined \def \showISBNx     #1{\unskip}     \fi
\ifx \showISBNxiii \undefined \def \showISBNxiii  #1{\unskip}     \fi
\ifx \showISSN     \undefined \def \showISSN      #1{\unskip}     \fi
\ifx \showLCCN     \undefined \def \showLCCN      #1{\unskip}     \fi
\ifx \shownote     \undefined \def \shownote      #1{#1}          \fi
\ifx \showarticletitle \undefined \def \showarticletitle #1{#1}   \fi
\ifx \showURL      \undefined \def \showURL       {\relax}        \fi
\providecommand\bibfield[2]{#2}
\providecommand\bibinfo[2]{#2}
\providecommand\natexlab[1]{#1}
\providecommand\showeprint[2][]{arXiv:#2}

\bibitem[\protect\citeauthoryear{Abel, Ara{\'u}jo, Gao, and Houben}{Abel
  et~al\mbox{.}}{2011}]%
        {abel2011analyzing}
\bibfield{author}{\bibinfo{person}{Fabian Abel}, \bibinfo{person}{Samur
  Ara{\'u}jo}, \bibinfo{person}{Qi Gao}, {and} \bibinfo{person}{Geert-Jan
  Houben}.} \bibinfo{year}{2011}\natexlab{}.
\newblock \showarticletitle{Analyzing cross-system user modeling on the social
  web}. In \bibinfo{booktitle}{\emph{International Conference on Web
  Engineering}}. Springer, \bibinfo{pages}{28--43}.
\newblock


\bibitem[\protect\citeauthoryear{Arjovsky and Bottou}{Arjovsky and
  Bottou}{2017}]%
        {arjovsky2017towards}
\bibfield{author}{\bibinfo{person}{Martin Arjovsky} {and}
  \bibinfo{person}{L{\'e}on Bottou}.} \bibinfo{year}{2017}\natexlab{}.
\newblock \showarticletitle{Towards principled methods for training generative
  adversarial networks}.
\newblock \bibinfo{journal}{\emph{arXiv preprint arXiv:1701.04862}}
  (\bibinfo{year}{2017}).
\newblock


\bibitem[\protect\citeauthoryear{Arjovsky, Chintala, and Bottou}{Arjovsky
  et~al\mbox{.}}{2017}]%
        {arjovsky2017wasserstein}
\bibfield{author}{\bibinfo{person}{Martin Arjovsky}, \bibinfo{person}{Soumith
  Chintala}, {and} \bibinfo{person}{L{\'e}on Bottou}.}
  \bibinfo{year}{2017}\natexlab{}.
\newblock \showarticletitle{Wasserstein gan}.
\newblock \bibinfo{journal}{\emph{arXiv preprint arXiv:1701.07875}}
  (\bibinfo{year}{2017}).
\newblock


\bibitem[\protect\citeauthoryear{Avazpour, Pitakrat, Grunske, and
  Grundy}{Avazpour et~al\mbox{.}}{2014}]%
        {avazpour2014dimensions}
\bibfield{author}{\bibinfo{person}{Iman Avazpour}, \bibinfo{person}{Teerat
  Pitakrat}, \bibinfo{person}{Lars Grunske}, {and} \bibinfo{person}{John
  Grundy}.} \bibinfo{year}{2014}\natexlab{}.
\newblock \showarticletitle{Dimensions and metrics for evaluating
  recommendation systems}.
\newblock In \bibinfo{booktitle}{\emph{Recommendation systems in software
  engineering}}. \bibinfo{publisher}{Springer}, \bibinfo{pages}{245--273}.
\newblock


\bibitem[\protect\citeauthoryear{Bharadhwaj, Park, and Lim}{Bharadhwaj
  et~al\mbox{.}}{2018}]%
        {bharadhwaj2018recgan}
\bibfield{author}{\bibinfo{person}{Homanga Bharadhwaj}, \bibinfo{person}{Homin
  Park}, {and} \bibinfo{person}{Brian~Y Lim}.} \bibinfo{year}{2018}\natexlab{}.
\newblock \showarticletitle{RecGAN: Recurrent generative adversarial networks
  for recommendation systems}. In \bibinfo{booktitle}{\emph{Proceedings of the
  12th ACM Conference on Recommender Systems}}. ACM, \bibinfo{pages}{372--376}.
\newblock


\bibitem[\protect\citeauthoryear{Blei, Ng, and Jordan}{Blei
  et~al\mbox{.}}{2003}]%
        {blei2003latent}
\bibfield{author}{\bibinfo{person}{David~M Blei}, \bibinfo{person}{Andrew~Y
  Ng}, {and} \bibinfo{person}{Michael~I Jordan}.}
  \bibinfo{year}{2003}\natexlab{}.
\newblock \showarticletitle{Latent dirichlet allocation}.
\newblock \bibinfo{journal}{\emph{Journal of machine Learning research}}
  \bibinfo{volume}{3}, \bibinfo{number}{Jan} (\bibinfo{year}{2003}),
  \bibinfo{pages}{993--1022}.
\newblock


\bibitem[\protect\citeauthoryear{Blondel, Ishihata, Fujino, and Ueda}{Blondel
  et~al\mbox{.}}{2016}]%
        {blondel2016polynomial}
\bibfield{author}{\bibinfo{person}{Mathieu Blondel}, \bibinfo{person}{Masakazu
  Ishihata}, \bibinfo{person}{Akinori Fujino}, {and} \bibinfo{person}{Naonori
  Ueda}.} \bibinfo{year}{2016}\natexlab{}.
\newblock \showarticletitle{Polynomial networks and factorization machines: New
  insights and efficient training algorithms}. In
  \bibinfo{booktitle}{\emph{Proceedings of International Conference on Machine
  Learning}}.
\newblock


\bibitem[\protect\citeauthoryear{Campos, Bellog{\'\i}n, D{\'\i}ez, and
  Chavarriaga}{Campos et~al\mbox{.}}{2010}]%
        {campos2010simple}
\bibfield{author}{\bibinfo{person}{Pedro~G Campos}, \bibinfo{person}{Alejandro
  Bellog{\'\i}n}, \bibinfo{person}{Fernando D{\'\i}ez}, {and}
  \bibinfo{person}{J~Enrique Chavarriaga}.} \bibinfo{year}{2010}\natexlab{}.
\newblock \showarticletitle{Simple time-biased KNN-based recommendations}. In
  \bibinfo{booktitle}{\emph{Proceedings of the Workshop on Context-Aware Movie
  Recommendation}}. ACM, \bibinfo{pages}{20--23}.
\newblock


\bibitem[\protect\citeauthoryear{Campos, D{\i}ez, and Cantador}{Campos
  et~al\mbox{.}}{2012}]%
        {campos2012performance}
\bibfield{author}{\bibinfo{person}{Pedro~G Campos}, \bibinfo{person}{Fernando
  D{\i}ez}, {and} \bibinfo{person}{Iv{\'a}n Cantador}.}
  \bibinfo{year}{2012}\natexlab{}.
\newblock \showarticletitle{A performance comparison of time-aware
  recommendation models}.
\newblock


\bibitem[\protect\citeauthoryear{Deng, Sang, Xu, et~al\mbox{.}}{Deng
  et~al\mbox{.}}{2013}]%
        {deng2013personalized}
\bibfield{author}{\bibinfo{person}{Zhengyu Deng}, \bibinfo{person}{Jitao Sang},
  \bibinfo{person}{Changsheng Xu}, {et~al\mbox{.}}}
  \bibinfo{year}{2013}\natexlab{}.
\newblock \showarticletitle{Personalized video recommendation based on
  cross-platform user modeling}. In \bibinfo{booktitle}{\emph{2013 IEEE
  International Conference on Multimedia and Expo (ICME)}}. IEEE,
  \bibinfo{pages}{1--6}.
\newblock


\bibitem[\protect\citeauthoryear{He, Chen, Kan, and Chen}{He
  et~al\mbox{.}}{2015}]%
        {he2015trirank}
\bibfield{author}{\bibinfo{person}{Xiangnan He}, \bibinfo{person}{Tao Chen},
  \bibinfo{person}{Min-Yen Kan}, {and} \bibinfo{person}{Xiao Chen}.}
  \bibinfo{year}{2015}\natexlab{}.
\newblock \showarticletitle{Trirank: Review-aware explainable recommendation by
  modeling aspects}. In \bibinfo{booktitle}{\emph{Proceedings of the 24th ACM
  International on Conference on Information and Knowledge Management}}. ACM,
  \bibinfo{pages}{1661--1670}.
\newblock


\bibitem[\protect\citeauthoryear{He, Liao, Zhang, Nie, Hu, and Chua}{He
  et~al\mbox{.}}{2017}]%
        {he2017neural}
\bibfield{author}{\bibinfo{person}{Xiangnan He}, \bibinfo{person}{Lizi Liao},
  \bibinfo{person}{Hanwang Zhang}, \bibinfo{person}{Liqiang Nie},
  \bibinfo{person}{Xia Hu}, {and} \bibinfo{person}{Tat-Seng Chua}.}
  \bibinfo{year}{2017}\natexlab{}.
\newblock \showarticletitle{Neural collaborative filtering}. In
  \bibinfo{booktitle}{\emph{Proceedings of the 26th International Conference on
  World Wide Web}}. International World Wide Web Conferences Steering
  Committee, \bibinfo{pages}{173--182}.
\newblock


\bibitem[\protect\citeauthoryear{Hidasi, Karatzoglou, Baltrunas, and
  Tikk}{Hidasi et~al\mbox{.}}{2015}]%
        {hidasi2015session}
\bibfield{author}{\bibinfo{person}{Bal{\'a}zs Hidasi},
  \bibinfo{person}{Alexandros Karatzoglou}, \bibinfo{person}{Linas Baltrunas},
  {and} \bibinfo{person}{Domonkos Tikk}.} \bibinfo{year}{2015}\natexlab{}.
\newblock \showarticletitle{Session-based recommendations with recurrent neural
  networks}.
\newblock \bibinfo{journal}{\emph{International Conference on Learning
  Representations}} (\bibinfo{year}{2015}).
\newblock


\bibitem[\protect\citeauthoryear{Kang, Fang, Wang, and McAuley}{Kang
  et~al\mbox{.}}{2017}]%
        {kang2017visually}
\bibfield{author}{\bibinfo{person}{Wang-Cheng Kang}, \bibinfo{person}{Chen
  Fang}, \bibinfo{person}{Zhaowen Wang}, {and} \bibinfo{person}{Julian
  McAuley}.} \bibinfo{year}{2017}\natexlab{}.
\newblock \showarticletitle{Visually-aware fashion recommendation and design
  with generative image models}. In \bibinfo{booktitle}{\emph{2017 IEEE
  International Conference on Data Mining (ICDM)}}. IEEE,
  \bibinfo{pages}{207--216}.
\newblock


\bibitem[\protect\citeauthoryear{Kingma and Ba}{Kingma and Ba}{2015}]%
        {kingma2014adam}
\bibfield{author}{\bibinfo{person}{Diederik~P Kingma} {and}
  \bibinfo{person}{Jimmy~Lei Ba}.} \bibinfo{year}{2015}\natexlab{}.
\newblock \showarticletitle{Adam: Amethod for stochastic optimization}. In
  \bibinfo{booktitle}{\emph{International Conference on Learning
  Representation}}.
\newblock


\bibitem[\protect\citeauthoryear{Lim, Lu, Chen, and Kan}{Lim
  et~al\mbox{.}}{2015}]%
        {lim2015mytweet}
\bibfield{author}{\bibinfo{person}{Bang~Hui Lim}, \bibinfo{person}{Dongyuan
  Lu}, \bibinfo{person}{Tao Chen}, {and} \bibinfo{person}{Min-Yen Kan}.}
  \bibinfo{year}{2015}\natexlab{}.
\newblock \showarticletitle{\# mytweet via instagram: Exploring user behaviour
  across multiple social networks}. In \bibinfo{booktitle}{\emph{Advances in
  Social Networks Analysis and Mining (ASONAM), 2015 IEEE/ACM International
  Conference on}}. IEEE, \bibinfo{pages}{113--120}.
\newblock


\bibitem[\protect\citeauthoryear{Mehta, Niederee, Stewart, Degemmis, Lops, and
  Semeraro}{Mehta et~al\mbox{.}}{2005}]%
        {mehta2005ontologically}
\bibfield{author}{\bibinfo{person}{Bhaskar Mehta}, \bibinfo{person}{Claudia
  Niederee}, \bibinfo{person}{Avare Stewart}, \bibinfo{person}{Marco Degemmis},
  \bibinfo{person}{Pasquale Lops}, {and} \bibinfo{person}{Giovanni Semeraro}.}
  \bibinfo{year}{2005}\natexlab{}.
\newblock \showarticletitle{Ontologically-enriched unified user modeling for
  cross-system personalization}. In \bibinfo{booktitle}{\emph{International
  Conference on User Modeling}}. Springer, \bibinfo{pages}{119--123}.
\newblock


\bibitem[\protect\citeauthoryear{Osborne, Petrovic, McCreadie, Macdonald, and
  Ounis}{Osborne et~al\mbox{.}}{2012}]%
        {osborne2012bieber}
\bibfield{author}{\bibinfo{person}{Miles Osborne}, \bibinfo{person}{Sa{\v{s}}a
  Petrovic}, \bibinfo{person}{Richard McCreadie}, \bibinfo{person}{Craig
  Macdonald}, {and} \bibinfo{person}{Iadh Ounis}.}
  \bibinfo{year}{2012}\natexlab{}.
\newblock \showarticletitle{Bieber no more: First story detection using Twitter
  and Wikipedia}. In \bibinfo{booktitle}{\emph{SIGIR 2012 Workshop on
  Time-aware Information Access}}.
\newblock


\bibitem[\protect\citeauthoryear{Ouyang, Liu, Rong, and Xiong}{Ouyang
  et~al\mbox{.}}{2014}]%
        {ouyang2014autoencoder}
\bibfield{author}{\bibinfo{person}{Yuanxin Ouyang}, \bibinfo{person}{Wenqi
  Liu}, \bibinfo{person}{Wenge Rong}, {and} \bibinfo{person}{Zhang Xiong}.}
  \bibinfo{year}{2014}\natexlab{}.
\newblock \showarticletitle{Autoencoder-based collaborative filtering}. In
  \bibinfo{booktitle}{\emph{International Conference on Neural Information
  Processing}}. Springer, \bibinfo{pages}{284--291}.
\newblock


\bibitem[\protect\citeauthoryear{Perera and Zimmermann}{Perera and
  Zimmermann}{2017}]%
        {perera2017exploring}
\bibfield{author}{\bibinfo{person}{Dilruk Perera} {and} \bibinfo{person}{Roger
  Zimmermann}.} \bibinfo{year}{2017}\natexlab{}.
\newblock \showarticletitle{Exploring the use of time-dependent cross-network
  information for personalized recommendations}. In
  \bibinfo{booktitle}{\emph{Proceedings of the 2017 ACM on Multimedia
  Conference}}. ACM, \bibinfo{pages}{1780--1788}.
\newblock


\bibitem[\protect\citeauthoryear{Perera and Zimmermann}{Perera and
  Zimmermann}{2018}]%
        {perera2018lstm}
\bibfield{author}{\bibinfo{person}{Dilruk Perera} {and} \bibinfo{person}{Roger
  Zimmermann}.} \bibinfo{year}{2018}\natexlab{}.
\newblock \showarticletitle{LSTM Networks for Online Cross-Network
  Recommendations.}. In \bibinfo{booktitle}{\emph{IJCAI}}.
  \bibinfo{pages}{3825--3833}.
\newblock


\bibitem[\protect\citeauthoryear{Radford, Metz, and Chintala}{Radford
  et~al\mbox{.}}{2015}]%
        {radford2015unsupervised}
\bibfield{author}{\bibinfo{person}{Alec Radford}, \bibinfo{person}{Luke Metz},
  {and} \bibinfo{person}{Soumith Chintala}.} \bibinfo{year}{2015}\natexlab{}.
\newblock \showarticletitle{Unsupervised representation learning with deep
  convolutional generative adversarial networks}.
\newblock \bibinfo{journal}{\emph{International Conference on Learning
  Representations (ICLR)}}.
\newblock


\bibitem[\protect\citeauthoryear{Rendle, Freudenthaler, Gantner, and
  Schmidt-Thieme}{Rendle et~al\mbox{.}}{2009}]%
        {rendle2009bpr}
\bibfield{author}{\bibinfo{person}{Steffen Rendle}, \bibinfo{person}{Christoph
  Freudenthaler}, \bibinfo{person}{Zeno Gantner}, {and} \bibinfo{person}{Lars
  Schmidt-Thieme}.} \bibinfo{year}{2009}\natexlab{}.
\newblock \showarticletitle{BPR: Bayesian personalized ranking from implicit
  feedback}. In \bibinfo{booktitle}{\emph{Proceedings of the twenty-fifth
  conference on uncertainty in artificial intelligence}}. AUAI Press,
  \bibinfo{pages}{452--461}.
\newblock


\bibitem[\protect\citeauthoryear{Roy, Mei, Zeng, and Li}{Roy
  et~al\mbox{.}}{2012}]%
        {roy2012socialtransfer}
\bibfield{author}{\bibinfo{person}{Suman~Deb Roy}, \bibinfo{person}{Tao Mei},
  \bibinfo{person}{Wenjun Zeng}, {and} \bibinfo{person}{Shipeng Li}.}
  \bibinfo{year}{2012}\natexlab{}.
\newblock \showarticletitle{Socialtransfer: Cross-domain transfer learning from
  social streams for media applications}. In
  \bibinfo{booktitle}{\emph{Proceedings of the 20th ACM International
  Conference on Multimedia}}. ACM, \bibinfo{pages}{649--658}.
\newblock


\bibitem[\protect\citeauthoryear{Sahoo, Singh, and Mukhopadhyay}{Sahoo
  et~al\mbox{.}}{2012}]%
        {sahoo2012hidden}
\bibfield{author}{\bibinfo{person}{Nachiketa Sahoo}, \bibinfo{person}{Param~Vir
  Singh}, {and} \bibinfo{person}{Tridas Mukhopadhyay}.}
  \bibinfo{year}{2012}\natexlab{}.
\newblock \showarticletitle{A hidden Markov model for collaborative filtering}.
\newblock \bibinfo{journal}{\emph{MIS Quarterly}} (\bibinfo{year}{2012}),
  \bibinfo{pages}{1329--1356}.
\newblock


\bibitem[\protect\citeauthoryear{Salakhutdinov, Mnih, and Hinton}{Salakhutdinov
  et~al\mbox{.}}{2007}]%
        {salakhutdinov2007restricted}
\bibfield{author}{\bibinfo{person}{Ruslan Salakhutdinov},
  \bibinfo{person}{Andriy Mnih}, {and} \bibinfo{person}{Geoffrey Hinton}.}
  \bibinfo{year}{2007}\natexlab{}.
\newblock \showarticletitle{Restricted Boltzmann machines for collaborative
  filtering}. In \bibinfo{booktitle}{\emph{Proceedings of the 24th
  International Conference on Machine learning}}. ACM,
  \bibinfo{pages}{791--798}.
\newblock


\bibitem[\protect\citeauthoryear{Salimans, Goodfellow, Zaremba, Cheung,
  Radford, and Chen}{Salimans et~al\mbox{.}}{2016}]%
        {salimans2016improved}
\bibfield{author}{\bibinfo{person}{Tim Salimans}, \bibinfo{person}{Ian
  Goodfellow}, \bibinfo{person}{Wojciech Zaremba}, \bibinfo{person}{Vicki
  Cheung}, \bibinfo{person}{Alec Radford}, {and} \bibinfo{person}{Xi Chen}.}
  \bibinfo{year}{2016}\natexlab{}.
\newblock \showarticletitle{Improved techniques for training gans}. In
  \bibinfo{booktitle}{\emph{Advances in Neural Information Processing
  Systems}}. \bibinfo{pages}{2234--2242}.
\newblock


\bibitem[\protect\citeauthoryear{Wang, Yu, Zhang, Gong, Xu, Wang, Zhang, and
  Zhang}{Wang et~al\mbox{.}}{2017}]%
        {wang2017irgan}
\bibfield{author}{\bibinfo{person}{Jun Wang}, \bibinfo{person}{Lantao Yu},
  \bibinfo{person}{Weinan Zhang}, \bibinfo{person}{Yu Gong},
  \bibinfo{person}{Yinghui Xu}, \bibinfo{person}{Benyou Wang},
  \bibinfo{person}{Peng Zhang}, {and} \bibinfo{person}{Dell Zhang}.}
  \bibinfo{year}{2017}\natexlab{}.
\newblock \showarticletitle{Irgan: A minimax game for unifying generative and
  discriminative information retrieval models}. In
  \bibinfo{booktitle}{\emph{Proceedings of the 40th International ACM SIGIR
  conference on Research and Development in Information Retrieval}}. ACM,
  \bibinfo{pages}{515--524}.
\newblock


\bibitem[\protect\citeauthoryear{Yan, Sang, and Xu}{Yan et~al\mbox{.}}{2014}]%
        {yan2014mining}
\bibfield{author}{\bibinfo{person}{Ming Yan}, \bibinfo{person}{Jitao Sang},
  {and} \bibinfo{person}{Changsheng Xu}.} \bibinfo{year}{2014}\natexlab{}.
\newblock \showarticletitle{Mining cross-network association for youtube video
  promotion}. In \bibinfo{booktitle}{\emph{Proceedings of the 22nd ACM
  International conference on Multimedia}}. ACM, \bibinfo{pages}{557--566}.
\newblock


\bibitem[\protect\citeauthoryear{Yoo, Ha, Yi, Ryu, Kim, Ha, Kim, and Yoon}{Yoo
  et~al\mbox{.}}{2017}]%
        {yoo2017energy}
\bibfield{author}{\bibinfo{person}{Jaeyoon Yoo}, \bibinfo{person}{Heonseok Ha},
  \bibinfo{person}{Jihun Yi}, \bibinfo{person}{Jongha Ryu},
  \bibinfo{person}{Chanju Kim}, \bibinfo{person}{Jung-Woo Ha},
  \bibinfo{person}{Young-Han Kim}, {and} \bibinfo{person}{Sungroh Yoon}.}
  \bibinfo{year}{2017}\natexlab{}.
\newblock \showarticletitle{Energy-based sequence gans for recommendation and
  their connection to imitation learning}.
\newblock \bibinfo{journal}{\emph{Proceedings of ACM Conference}}.
\newblock


\bibitem[\protect\citeauthoryear{Zhang}{Zhang}{2013}]%
        {zhang2013definition}
\bibfield{author}{\bibinfo{person}{Liang Zhang}.}
  \bibinfo{year}{2013}\natexlab{}.
\newblock \showarticletitle{The definition of novelty in recommendation
  system.}
\newblock \bibinfo{journal}{\emph{Journal of Engineering Science \& Technology
  Review}} \bibinfo{volume}{6}, \bibinfo{number}{3}.
\newblock


\bibitem[\protect\citeauthoryear{Zhao, Jiang, Weng, He, Lim, Yan, and Li}{Zhao
  et~al\mbox{.}}{2011}]%
        {zhao2011comparing}
\bibfield{author}{\bibinfo{person}{Wayne~Xin Zhao}, \bibinfo{person}{Jing
  Jiang}, \bibinfo{person}{Jianshu Weng}, \bibinfo{person}{Jing He},
  \bibinfo{person}{Ee-Peng Lim}, \bibinfo{person}{Hongfei Yan}, {and}
  \bibinfo{person}{Xiaoming Li}.} \bibinfo{year}{2011}\natexlab{}.
\newblock \showarticletitle{Comparing twitter and traditional media using topic
  models}. In \bibinfo{booktitle}{\emph{European Conference on Information
  Retrieval}}. Springer, \bibinfo{pages}{338--349}.
\newblock


\end{thebibliography}
